\documentclass[10pt,twocolumn,letterpaper]{article}

\usepackage{iccv}
\usepackage{times}
\usepackage{epsfig}
\usepackage{graphicx}
\usepackage{amsmath}
\usepackage{amssymb}
\usepackage{booktabs}
\usepackage{subcaption}

\usepackage[pagebackref=true,breaklinks=true,letterpaper=true,colorlinks,bookmarks=false]{hyperref}

\usepackage{multirow}
\usepackage{booktabs}
\usepackage{enumitem}

\def\eg{\emph{e.g., }}
\def\ie{\emph{i.e., }}

\def\wrt{\emph{w.r.t. }}

\def\tab{Table }
\def\fig{Figure }
\def\equ{Equation }
\def\sec{Section }
\def\name{STPrivacy}
\def\tubs{tubelets}
\def\tub{tubelet}

\def\spars{sparsification}

\def\anony{anonymization}

\DeclareMathOperator{\MLP}{MLP}
\DeclareMathOperator{\Avg}{Avg}
\DeclareMathOperator{\Flatten}{Flat}
\DeclareMathOperator{\Expd}{Expd}
\DeclareMathOperator{\Concat}{Concat}
\DeclareMathOperator{\softmax}{Softmax}
\DeclareMathOperator{\gumbelsoftmax}{GumbelSoftmax}

\usepackage{pifont}

\usepackage{xcolor}
\definecolor{britishracinggreen}{rgb}{0.0, 0.26, 0.15}
\definecolor{aoenglish}{rgb}{0.0, 0.5, 0.0}

\usepackage{mathtools}

\makeatletter
\newcommand*{\rom}[1]{\expandafter\@slowromancap\romannumeral #1@}
\makeatother
\makeatletter
\newcommand\footnoteref[1]{\protected@xdef\@thefnmark{\ref{#1}}\@footnotemark}
\makeatother
\newcommand{\bfsection}[1]{\vspace*{0.5mm}\noindent\textbf{#1.}}

\iccvfinalcopy 

\ificcvfinal\fi

\begin{document}

\title{\name{}: Spatio-Temporal Privacy-Preserving Action Recognition}

\author{
Ming Li\textsuperscript{\rm 1} \quad
Xiangyu Xu\textsuperscript{\rm 2} \quad 
Hehe Fan\textsuperscript{\rm 4} \quad
Pan Zhou\textsuperscript{\rm 2} \quad
Jun Liu\textsuperscript{\rm 3} \quad
Jia-Wei Liu\textsuperscript{\rm 5} \\
Jiahe Li\textsuperscript{\rm 4} \quad
Jussi Keppo\textsuperscript{\rm 6} \quad
Mike Zheng Shou\textsuperscript{\rm 5} \quad
Shuicheng Yan\textsuperscript{\rm 2} \\
\textsuperscript{\rm 1}Institute of Data Science, National University of Singapore \\
\textsuperscript{\rm 2} Sea AI Lab\\
\textsuperscript{\rm 3}Singapore University of Technology and Design \\
\textsuperscript{\rm 4}School of Computing, National University of Singapore \\
\textsuperscript{\rm 5} Show Lab, National University of Singapore \\
\textsuperscript{\rm 6} Business School, National University of Singapore
\\
}

\maketitle
\ificcvfinal\thispagestyle{empty}\fi

\begin{abstract}
   Existing methods of privacy-preserving action recognition (PPAR) mainly focus on frame-level (\emph{spatial}) privacy removal through 2D CNNs. Unfortunately, they have two major drawbacks. First, they may compromise temporal dynamics in input videos, which are critical for accurate action recognition. Second, they are vulnerable to practical attacking scenarios where attackers probe for privacy from an entire video rather than individual frames. To address these issues, we propose a novel framework \name{} to perform video-level PPAR. For the first time, we introduce vision Transformers into PPAR by treating a video as a \tub{} sequence, and accordingly design two complementary mechanisms, \ie \spars{} and \anony{}, to remove privacy from a \emph{spatio-temporal} perspective. In specific, our privacy \spars{} mechanism applies adaptive token selection to abandon action-irrelevant \tubs{}. Then, our \anony{} mechanism implicitly manipulates the remaining action-\tubs{} to erase privacy in the embedding space through adversarial learning. These mechanisms provide significant advantages in terms of privacy preservation for human eyes and action-privacy trade-off adjustment during deployment. We additionally contribute \emph{the first two} large-scale PPAR benchmarks, VP-HMDB51 and VP-UCF101, to the community. Extensive evaluations on them, as well as two other tasks, validate the effectiveness and generalization capability of our framework.
\end{abstract}

\begin{figure}[!h]
\setlength\abovecaptionskip{3mm}
\centering
\includegraphics[width=0.99\columnwidth]{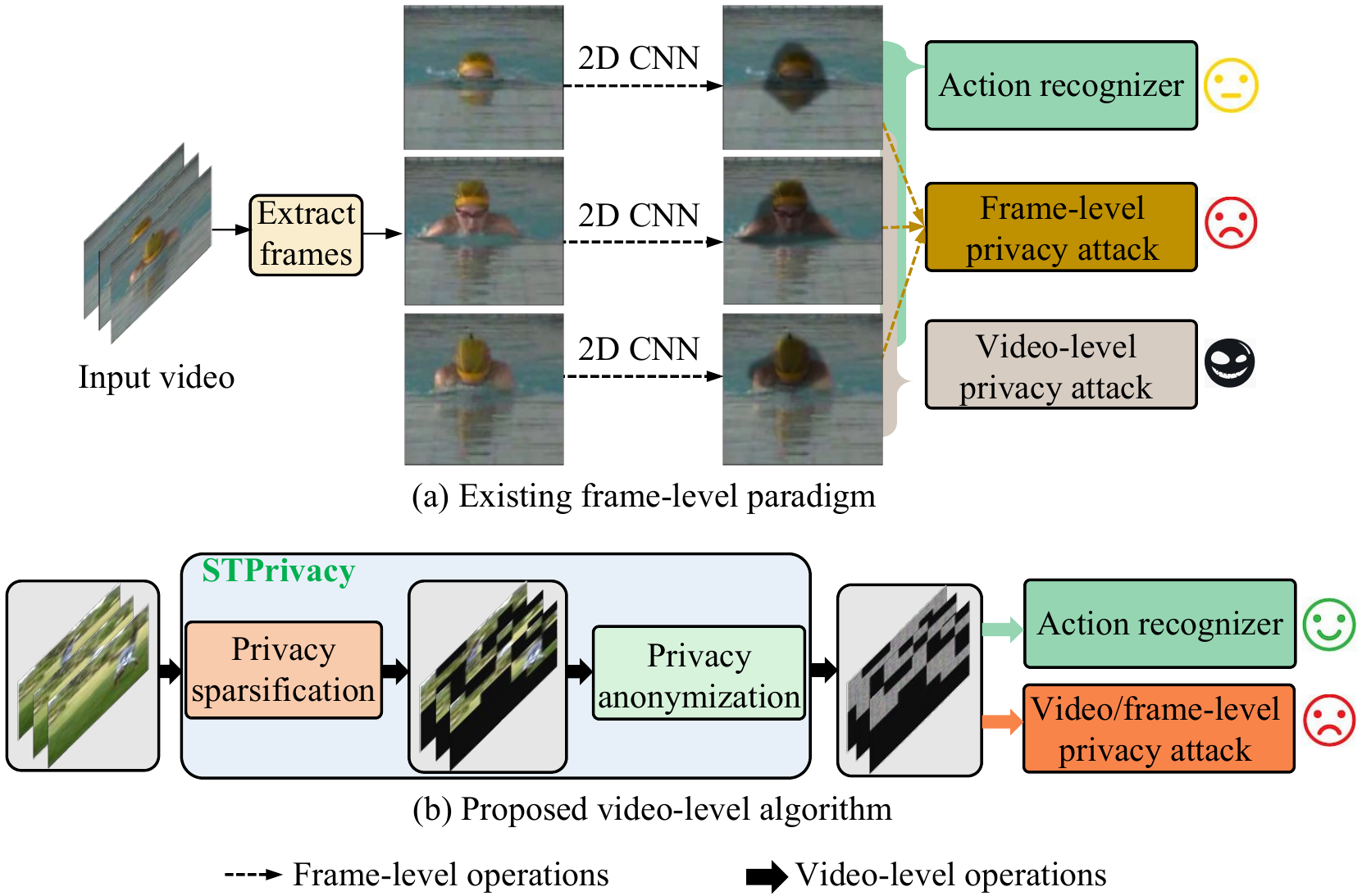}
\vspace{-2mm}
\caption{Comparison between the existing paradigm for PPAR and the proposed algorithm. (a) Existing methods for PPAR remove private information from individual frames \textit{independently} against a \textit{frame-level} privacy recognizer. They not only neglect the temporal dynamics between frames, hurting action recognition performance, but also leave the entire video vulnerable to privacy attacks. (b) Our proposed algorithm addresses these issues by treating the input video as a whole to remove privacy against a \textit{video-level} privacy recognizer. It promotes action dynamics and protects both video-level and frame-level privacy. The black rectangles in our transformed video represent the abandoned \tubs{}, and the emotional faces indicate the performance of the corresponding task.}
\vspace{-5mm}
\label{fig:existing_ours}
\end{figure}

\section{Introduction}
\label{sec:intro}
 Action recognition has seen tremendous progress in recent years \cite{kong2022human}, but the increasing concerns regarding privacy leakage have given rise to an emerging research topic privacy-preserving action recognition (PPAR) \cite{wu_eccv, wu_tpami, dave2022spact}. It aims to remove private information from videos while ensuring accurate action recognition.

Current studies on PPAR \cite{srivastav2019human, liu2020indoor,zhang2021multi, ren2018learning, wu_eccv, wu_tpami, dave2022spact} mainly focus on frame-level privacy preservation. As illustrated in \fig \ref{fig:existing_ours} (a), the paradigm typically involves three steps: 1) extracting frames from a video, 2) independently removing privacy from each frame, and 3) performing privacy recognition and action recognition on the transformed frames and the spliced pseudo video, respectively. 

While this paradigm is effective against frame-level privacy attacks, it has two major drawbacks. First, it neglects the temporal dynamics between frames, which are crucial for accurate action recognition \cite{tsn_journal, slowfast}. This is because it usually relies on a 2D convolutional neural network (CNN) to process each video frame independently, resulting in a serious discontinuity in object dynamics. Second, the paradigm only protects spatial privacy against frame-level privacy attacks, leaving the entire video vulnerable to potential video-level privacy attacks. A typical example is that it can merely remove part areas of a face to make it difficult to solely identify from each individual frame. But a video-level privacy recognizer can still identify the face by aggregating the highly complementary facial clues from the remaining areas of all frames, owing to the high information redundancy in a video \cite{tong2022videomae, feichtenhofer2022masked}. The essential techniques can be obtained referring to the research on occluded video object recognition~\cite{hou2021feature, Wang_2021_ICCV, hou2019vrstc, zhang2020multi}.

To overcome these drawbacks of frame-level PPAR, we present a novel algorithm, named \name{}, which performs video-level PPAR from both spatial and temporal perspectives as illustrated in \fig\ref{fig:existing_ours} (b). 
Inspired by the latest vision Transformers (ViTs), \name{} treats an input video as a \tub{} sequence and captures the temporal dynamics with the self-attention operations. To enable the privacy removal within our framework, we propose two complementary token-wise mechanisms, namely \emph{\spars{}} and \emph{\anony{}}. The privacy \spars{} mechanism applies adaptive token selection to directly abandon the private \tubs{} that are irrelevant to the action. Then, the privacy \anony{} mechanism manipulates the remaining action-\tubs{} in the embedding space to implicitly erase privacy. For training the proposed network, we employ an adversarial learning objective by feeding the transformed \tubs{} into an action recognizer and a video-level privacy recognizer. Both intuitively and experimentally, our \name{} is superior in protecting both video-level (\sec \ref{sec:sota}) and frame-level (\sec \ref{sec:frame-level-test}) privacy. Clearly, our framework emphasizes temporal dynamics for action recognition and protects privacy in a more strict manner. 

In summary, our main contributions are as follows:
\begin{itemize}[nosep,]
\item We propose a novel video-level PPAR framework that enhances temporal dynamics for action recognition and protects privacy in a more strict way, compared with existing frame-level methods.

\item The proposed \name{} introduces ViTs into PPAR for the first time and demonstrates significant advantages in high-quality privacy preservation in terms of human eyes (Section \ref{sec:qualitative_analysis}) and convenient adjustment of action-privacy recognition trade-off during deployment (Section \ref{sec:ablation}). 

\item We provide new benchmark datasets VP-HMDB51 and VP-UCF101, which are considerably larger than the existing one (\ie PA-HMDB \cite{wu_tpami} containing 515 videos), for evaluating PPAR methods sufficiently. Our annotations will be made publicly available.

\item Extensive experiments demonstrate that \name{} significantly outperforms the state-of-the-art (SOTA) methods in terms of action recognition and privacy protection, both quantitatively and qualitatively. In addition, its generalization ability is demonstrated by two related tasks on CelebVHQ~\cite{zhu2022celebvhq} and P-HVU~\cite{dave2022spact}.
\end{itemize}

\begin{figure*}[t]
\setlength\abovecaptionskip{1mm}
\centering
\includegraphics[width=0.85\textwidth]{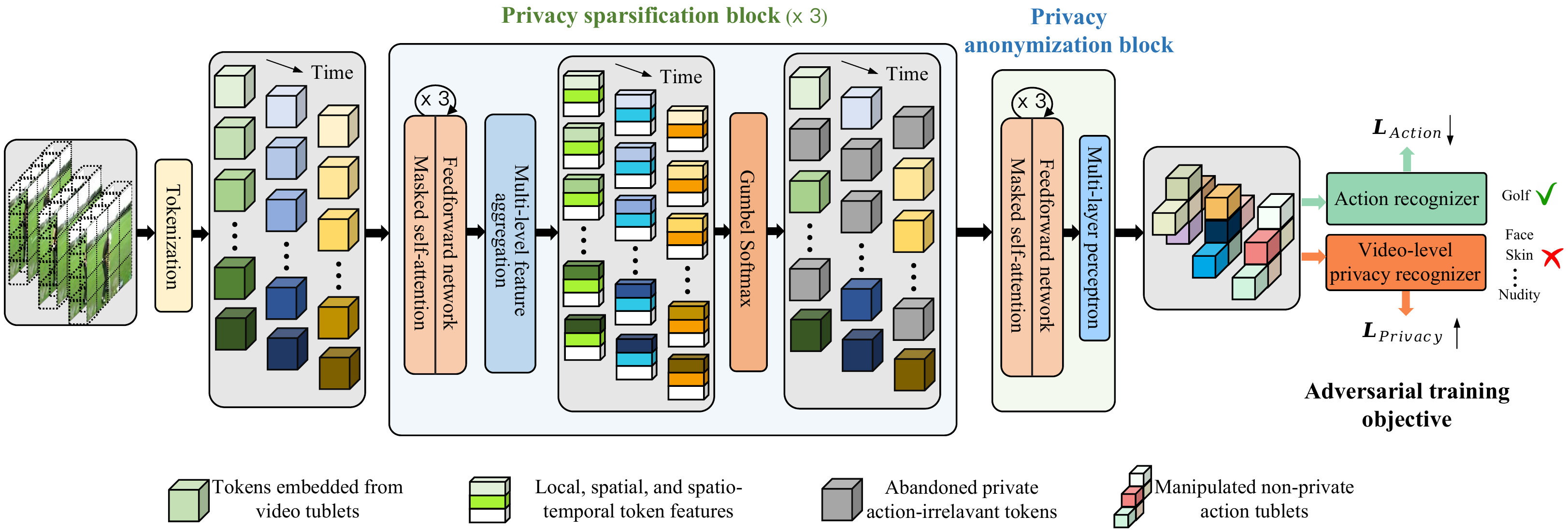}
\caption{Overview of the proposed \name{}, which aims to maintain action clues while removing private information during the transformation of raw videos. Its effectiveness is demonstrated by the stable performance of an action recognizer on the transformed videos, in contrast to the severe degradation experienced by a video-level privacy recognizer. Both the action and video-level privacy recognizers employed are regular ViT classifiers, which serve as auxiliary components for deriving an adversarial training objective. The former employs a common cross-entropy (CE) loss for supervising action recognition ({\small$\mathcal{L}_{\rm Action}$}), while the latter utilizes a multi-label binary CE loss for supervising privacy recognition ({\small $\mathcal{L}_{\rm Privacy}$}).
}
\vspace{-5mm}
\label{fig:overview}
\end{figure*}

\section{Related works} \label{sec:related_works}
\subsection{Privacy-preserving action recognition}
The existing literature in this field primarily focuses on frame-level privacy preservation. Researchers have categorized these efforts into three main streams based on their privacy-removal strategies: 1) spatial downsampling \cite{chou2018privacy, srivastav2019human, butler2015privacy, ryoo, ishwar, liu2020indoor}, 2) private area modification with hand-crafted operations \cite{zhang2021multi, ren2018learning}, and 3) learning-based transformation \cite{wu_eccv, wu_tpami, dave2022spact}. Spatial downsampling treats private and non-private areas of frames equally, which severely hinders action recognition when removing privacy. Private area modification relies on a pre-trained object detector to identify sensitive regions, which are then modified using pre-defined operations. 
However, this is an offline privacy-removal manner whose performance is dependent on domain shifts between the training data of the detector and the target data. Moreover, it only alters the detected areas, leading to severe data distribution gaps within a frame \cite{zhang2021multi, ren2018learning}. Learning-based transformation is a more promising strategy for balancing action-privacy recognition trade-offs \cite{wu_eccv, wu_tpami, dave2022spact}. However, current research in this direction mainly concentrates on privacy removal of individual frames. In this work, we present a novel video-level PPAR approach that benefits object dynamics and enforces more stringent privacy protection.

\subsection{Vision Transformer} \label{sec:related_works_vit}
Transformers with self-attention mechanisms \cite{attentions} have made tremendous progress in modeling deep correlations over long distances in natural languages \cite{informer, devlin2018bert}. ViT \cite{vit} is a notable instance of applying Transformers in image recognition, achieved by dividing an image into a sequence of patches and extracting embedded tokens from each patch. To address the challenges of video understanding, recent research, such as ViViT \cite{arnab2021vivit} and Timesformer \cite{bertasius2021space}, has explored various self-attention factorization techniques to capture spatio-temporal interactions. Additionally, a few works have recently proposed efficient techniques for performing ViT inference on image tasks \cite{rao2021dynamicvit, pan2021ia, kong2021spvit, meng2022adavit, yin2022vit}. Our framework significantly distinguishes itself from these efficient ViTs based on at least three aspects. Firstly, they focus on image recognition and do not model temporal dynamics, whereas our framework deals with videos and incorporates specially designed strategies for spatio-temporal information aggregation.  
Secondly, the discarded tokens in the efficient ViTs are implicitly involved in the final image class prediction because their information has been compressed into the package tokens or class tokens before being discarded~\cite{rao2021dynamicvit, pan2021ia, kong2021spvit, meng2022adavit, yin2022vit}, which inevitably leads to severe privacy leakage. Hence, these models can not be applied to PPAR. In contrast, our framework does not suffer from these drawbacks.

Finally, training the efficient ViTs typically requires the carefully designed guidance from a vanilla teacher to supervise the learning of their prediction logits or even token embeddings~\cite{rao2021dynamicvit, kong2021spvit, yin2022vit}, while our learning procedure does not rely on any external guidance.

\section{Methodology}
The overview of our proposed \name{} is illustrated in \fig \ref{fig:overview}. It consists of three sequential privacy \spars{} blocks (PSBs) and a privacy \anony{} block (PAB). The PSBs are designed to sparsify privacy in a video by adaptively abandoning the private action-irrelevant \tubs{}. Then the PAB, on the other hand, is responsible for manipulating the remaining action-\tubs{} in the embedding space to further remove privacy. Finally, the transformed \tubs{} are fed into an auxiliary action recognizer and an auxiliary video-level privacy recognizer, deriving an adversarial learning objective. The objective is employed to supervise the \name{} learning, minimizing the action recognition loss while maximizing the privacy recognition loss. 

The visual effect of our privacy \spars{} and \anony{} mechanisms is illustrated in \fig \ref{fig:existing_ours} (b). It is worth noting that prior learning-based methods \cite{wu_eccv, wu_tpami, dave2022spact} mainly focus on the embedding \anony{} to remove privacy, which can be seen as a special case of our \name{} where none of the privacy-containing \tubs{} are abandoned.

\subsection{Video tokenization} 
Let {\small$\{({\boldsymbol v},{\boldsymbol y}, {\boldsymbol p})\}$} denote a training dataset, where {\small${\boldsymbol v}\in\mathbb{R}^{T\times H\times W\times 3}$} is a video with height $H$, width $W$, and temporal length $T$.
{\small${\boldsymbol y} \in \{0,1\}^C$} is the one-hot label of the video over $C$ action classes. 
{\small${\boldsymbol p} \in \{0,1\}^P$} represents $P$ binary privacy labels, where the $i$-th entry of {\small${\boldsymbol p}$} indicates whether the $i$-th privacy attribute (face, skin, etc.) is exposed in the input video {\small${\boldsymbol v}$}.

To apply our Transformer-based framework, we convert the input video into a sequence of tokens {\small${\boldsymbol x}\in\mathbb{R}^{L\times N\times D}$}, where each token is a $D$-dimensional feature vector, extracted from a video \tub{} with the size {\small $\delta T\times\delta H\times\delta W\times 3$} using 3D convolutions.
All \tubs{} of ${\boldsymbol v}$ are non-overlapping, and each tubelet exactly corresponds to one token. Hence, we have $L=T/\delta T, N=H/\delta H \cdot W/\delta W$, where the spatial dimensions are flattened.
Additionally, we maintain a binary decision matrix ${\hat{\mathbf{I}}}\in \{0, 1\}^{L\times N}$ with all elements initialized as 1 to indicate whether a token of {\small${\boldsymbol x}$} is abandoned (0) or retained (1) during privacy sparsification. 

\subsection{Privacy \spars{}}
\label{sec:tube_sparse}
The \spars{} mechanism of \name{} is devised to adaptively remove \tubs{} that are private and do not contribute to action dynamics from a raw video. 
Specifically, for each PSB in Figure~\ref{fig:overview}, we first apply stacked Transformer layers, including a masked self-attention and a feedforward network, to learn evolved feature representations for the input tokens.
We then introduce a multi-level feature aggregation module to incorporate multi-scope global information into these tokens, which are subsequently used to predict the retaining probability of each video \tub{}.

\bfsection{Multi-level feature aggregation}
To take comprehensive clues into consideration when deciding the retaining probability of each token, we propose to perform multi-level feature aggregation, collecting local, spatial and spatio-temporal information into each token. In detail, a multi-layer perceptron (MLP) consisting of one linear layer followed by GELU activation \cite{hendrycks2016gaussian} is applied to map the input tokens $\boldsymbol x$ as the local feature:
\vspace{-1mm}
\begin{equation}
    {\boldsymbol x}^{\rm local} = \MLP({\boldsymbol x})\in \mathbb{R}^{L\times N \times D/3}.\vspace{-1mm}
\end{equation}
Next, we apply another MLP to obtain the spatial feature:
\vspace{-1mm}
\begin{equation}
    {\boldsymbol x}^{\rm spatio} = \Expd_{\rm s}(\Avg_{\rm s}(\MLP({\boldsymbol x}), \hat{\mathbf{I}}))\in \mathbb{R}^{L\times N \times D/3},\vspace{-2mm}
\end{equation}
where $\Avg$ and $\Expd$ represent averaging a 3D tensor and then expanding it by repeating, and the subscript ``{\rm s}'' indicates that the computations are conducted along the spatial dimension. 
Note that $\Avg$ is conditioned on the current decision matrix $\hat{\mathbf{I}}$ as only the remaining tokens in $\hat{\mathbf{I}}$ are averaged. 
Similarly, we can obtain the spatio-temporal feature: 
\vspace{-1mm}
{\small
\begin{equation}
    {\boldsymbol x}^{\rm spatem} = \Expd_{\rm st}(\Avg_{\rm st}(\MLP({\boldsymbol x}), \hat{\mathbf{I}}))\in \mathbb{R}^{L\times N \times D/3},\vspace{0mm}
\end{equation}}
where $\Avg_{\rm st}$ and $\Expd_{\rm st}$ represent the averaging and expanding operations over the spatio-temporal dimensions. Then these hierarchical features are concatenated along the last dimension as the \spars{} evidence of each token:
\vspace{-2mm}
\begin{equation}
    {\boldsymbol x}^{\rm spars} = \Concat({\boldsymbol x}^{\rm local}, {\boldsymbol x}^{\rm spatio}, {\boldsymbol x}^{\rm spatem}).\vspace{-1.5mm}
\end{equation}

\bfsection{Progressive token pruning}
With the aggregated multi-level features ${\boldsymbol x}^{\rm spars}$,
we use a three-layer MLP followed by a $\softmax$ operator to predict the token-retaining probabilities ${\boldsymbol z}$:
\vspace{-1.5mm}
\begin{equation}
    {\boldsymbol z} = \softmax(\MLP({\boldsymbol x}^{\rm spars}))\in \mathbb{R}^{L\times N \times 2}.\vspace{-1mm}
\end{equation}
Then we can sparsify the video privacy by pruning tokens according to the predicted probability ${\boldsymbol z}$.
However, normal sampling operations are non-differentiable with respect to the probability distribution, which makes it infeasible to train our framework in an end-to-end manner.
To circumvent this issue, we apply the Gumbel-Softmax~\cite{jang2016categorical, kong2021spvit, rao2021dynamicvit} for our differentiable \spars{}:
\vspace{-1mm}
\begin{equation}
    \mathbf{I} = \gumbelsoftmax({\boldsymbol z})\in \{0, 1\}^{L\times N}.\vspace{-1mm}
\end{equation}
The decision matrix $\hat{\mathbf{I}}$ is updated by its Hadamard product $\hat{\mathbf{I}}=\hat{\mathbf{I}}\odot\mathbf{I}$ and then used in subsequent computations. 

Instead of sparsifying the video privacy all in one step, we introduce a progressive sparsification schedule to better identify action-irrelevant private video \tubs{}, where we successively apply three PSBs as shown in Figure~\ref{fig:overview}. To enable stable \tub{} \spars{}, we keep a proportion $\alpha=0.7$ of the feeding tokens in each PSB by default. Besides, we find that simply applying this constraint across spatio-temporal dimensions easily causes training instability. Therefore, we encourage it on the spatial dimension of $\hat{\mathbf{I}}$ with a mean squared error loss: 
\vspace{-2mm}
\begin{equation} \label{equ:loss_ratio}
{\small
    \mathcal{L}_{\rm Spars}={\frac1{ML}}\sum_{m=1}^M\sum_{l=1}^L({\frac1N}\sum_{n=1}^N{\hat{\mathbf I}}_{(m)}{(l,n)-\alpha^m)}^2,
    }\vspace{-2mm}
\end{equation}
where $M=3$ is the total number of PSBs and ${\hat{\mathbf I}}_{(m)}$ represents the decision matrix of the m-$th$ PSB.

\bfsection{Masked self-attention} 
For regular ViTs, self-attention is naturally computed among all tokens to model their interactions \cite{attentions}. 
In the proposed \name{}, however, private action-irrelevant tokens are progressively pruned, and we only need to model the interactions among the remaining tokens \cite{kong2021spvit, rao2021dynamicvit}. In this case, the regular computation technique for self-attention is not feasible.
Therefore, we introduce a masked self-attention that better suits our task in this work.
A key problem of the masked self-attention is that the numbers of remaining tokens of videos in a training batch may differ. 
To address this issue, we use a mask matrix $\mathbf{M}$ to limit the information exchanging scope, and the attention $\tilde{\mathbf{A}}_{ij}$ can be computed as follows:
\vspace{-1mm}
\begin{align}
    &\mathbf{S}=\mathbf{Q}\mathbf{K}^{\rm T}/\sqrt{D} \in \mathbb{R}^{LN\times LN},\\
    &\mathbf{M}_{i,j} = 
    \begin{cases}
    1,&i=j,\\
    \Flatten_{\rm st}(\hat{\mathbf{I}})_j,& i\neq j.
    \end{cases}& 1\le i,j\le LN,\label{equ:W_ij}\\
    &\tilde{\mathbf{A}}_{ij}=  \frac{\exp(\mathbf{S}_{ij})\mathbf{M}_{ij}}{\sum_{k=1}^{LN}\exp(\mathbf{S}_{ik})\mathbf{M}_{ik}},& 1\le i,j\le LN.\label{equ:attention_masking}
\end{align}
{\small $\mathbf{Q}$, $\mathbf{K}\in \mathbb{R}^{LN \times D}$} are the projected query and key matrices respectively from $\Flatten_{\rm st}(\boldsymbol x)$, where $\Flatten_{\rm st}$ represents flattening the spatio-temporal dimensions of $\boldsymbol x$. During inference, it is straightforward to select a proportion $\alpha^3$ of all tokens for each video by sorting their retaining probabilities ${\boldsymbol z}$. This ensures that the same number of tokens are retained for videos in a batch, allowing for the computation of the self-attention in a regular manner.

\subsection{Privacy \anony{}}
\label{sec:tub_anony} 
A proportion of video \tubs{} not only contain rich private information but also play critical roles in representing action dynamics. Abandoning these \tubs{} can severely harm the performance of action recognition. To address this issue, we introduce the privacy \anony{} mechanism to manipulate \tub{} embeddings, implicitly removing private information from the remaining action-\tubs{} selected by the \spars{} mechanism. 

In prior studies \cite{wu_eccv, wu_tpami, dave2022spact}, complex networks such as UNet \cite{unet} and ITNet \cite{johnson2016perceptual} have been employed to manipulate frame embeddings for \anony{}. In contrast, we introduce a simple yet effective PAB consisting of three Transformer layers and a single-layer MLP for this purpose. The anonymized output can be obtained by inputting the tokens ${\boldsymbol x}$ into the MLP of PAB:
\vspace{-1mm}
\begin{equation}
    {\boldsymbol x}^{\rm anony} = \MLP({\boldsymbol x})\in \mathbb{R}^{LN\times3\delta T\delta H\delta W}.\vspace{-1mm}
\end{equation}
Finally, ${\boldsymbol x}^{\rm anony}$ is reshaped to the size of {\small$T\times H\times W\times 3$}, representing the transformed version of the raw video ${\boldsymbol v}$.

\subsection{Optimization procedure and objectives}
The process of performing PPAR typically comprises three phases \cite{dave2022spact}: framework initialization, adversarial learning, and evaluation.

\bfsection{Initialization}
We initialize \name{} on a PPAR benchmark dataset by adding a classification loss $\mathcal{L}^{'}_{\rm Action}$, which is similar to $\mathcal{L}_{\rm Action}$, to perform action recognition by itself. The full objective is: 
\vspace{-2mm}
\begin{equation}
    \mathcal{L}_{\rm }=\mathcal{L}_{\rm Spars} + \mathcal{L}^{'}_{\rm Action}.\vspace{-2mm}
\end{equation}  

\bfsection{Adversarial learning}
We employ an auxiliary action recognizer and an auxiliary video-privacy recognizer to derive an adversarial learning objective for updating \name{}:
\vspace{-2mm}
\begin{equation} \label{equ:sa}
    \mathcal{L}_{\rm }=\mathcal{L}_{\rm Spars} + \lambda_{\rm Action}\mathcal{L}_{\rm Action} - \lambda_{\rm Privacy}\mathcal{L}_{\rm Privacy}, \vspace{-2mm}
\end{equation}
where $\lambda_{\rm Action}$ and $\lambda_{\rm Privacy}$ are the weighting coefficients set to 0.5 by default. Iteratively, we update the action recognizer and video-privacy recognizer using $\mathcal{L}_{\rm Action}$ and $\mathcal{L}_{\rm Privacy}$, respectively, along with updating STPrivacy.

\bfsection{Evaluation}
We freeze the optimized \name{} and employ it to transform the raw videos. We then train an action recognizer and a video-privacy recognizer from scratch on the training split of the transformed videos. Finally, the trained recognizers are tested on the testing split to evaluate the effectiveness of our \name{}.

\section{Experiments}
\subsection{VP-HMDB and VP-UCF101 datasets}
The existing PPAR benchmark, \ie PA-HMDB~\cite{wu_tpami}, consists of only 515 videos, which may be not sufficient for evaluating a deep learning approach. Therefore, we collect two larger datasets, based on two of the most popular action recognition datasets. The first one is HMDB51 \cite{kuehne2011hmdb}, which contains 6,849 videos and 51 human actions. The second dataset is UCF101 \cite{soomro2012ucf101}, which contains 13,320 videos and 101 sports actions.

In line with the privacy definition used in previous studies \cite{wu_tpami, dave2022spact}, we annotated these two datasets using five human attributes, \ie \textit{face}, \textit{skin color}, \textit{gender}, \textit{nudity}, and \textit{familiar relationship}. We invited more than three annotators to meticulously review each video and independently determine a binary label for each attribute based on whether it is identifiable throughout the entire video. We applied majority voting to determine each label. We name these two benchmarks \textbf{VP-HMDB51} and \textbf{VP-UCF101}, respectively, and have adhered to the official training and testing split. 

We mainly conduct experiments on the newly collected VP-HMDB and VP-UCF101 datasets. The experimental results on the existing PA-HMDB are shown in \sec \ref{sec:frame-level-test}. 

\subsection{Evaluation metrics}

The performance of action recognition is evaluated by the widely used top-1 accuracy, where the class of a video is determined by the averaged prediction of 5 clips $\times$ 3 crops \cite{tong2022videomae}. Privacy recognition, on the other hand, is a multi-label binary classification problem. Its performance is evaluated using class-wise mean average precision (cMAP) and class-wise F1 score. These metrics are reported in percentages (top-1 and cMAP) and decimals (F1), where $\uparrow$ and $\downarrow$ represent that higher and lower values are better, respectively. The best values are marked in bold.

\begin{table}
\setlength{\tabcolsep}{2.5pt}
\centering
\resizebox{0.99\columnwidth}{!}{
\begin{tabular}{cccccccccc} 
\toprule
\multirow{2}{*}{Method} & \multicolumn{3}{c}{VP-HMDB51} & & \multicolumn{3}{c}{VP-UCF101}\\ 
\cmidrule{2-4}\cmidrule{6-8}
& Top-1 ($\uparrow$) & F1 ($\downarrow$)   & cMAP ($\downarrow$) & & Top-1 ($\uparrow$)  & F1 ($\downarrow$) & cMAP ($\downarrow$)\\ 
\midrule
Raw data & 51.44 & 0.673 & 75.58 & & 84.20 & 0.684 & 76.62\\
\midrule
Downsample-$2\times$ &40.80 &0.601 &71.35 &&72.79 &0.620 &71.49 \\
Downsample-$4\times$ &31.32 &0.594 &69.79  &&56.07 &0.615 &69.85 \\
\midrule
Blackening \cite{dave2022spact} &38.27 &0.649 &74.06  &&69.41  &0.660 &75.37\\
StrongBlur \cite{dave2022spact} &40.91 &0.655 &74.33  &&73.94  &0.672 &75.58\\
WeakBlur \cite{dave2022spact} &47.24 &0.663 &75.11  &&77.31  &0.678 &76.03\\
Collective \cite{icdm19} &46.88 &0.651 &74.12  &&78.01  &0.663 & 75.22\\
VITA \cite{wu_tpami} &48.11 &0.638 &73.89  &&78.49  &0.657 &75.36 \\
SPAct \cite{dave2022spact} &48.56 &0.642 &73.78  &&78.40 &0.651 &75.29\\
\midrule
\textbf{Ours} & \textbf{50.73} & \textbf{0.613} & \textbf{72.48}  &  & \textbf{82.55} & \textbf{0.634} & \textbf{73.79} \\
\bottomrule
\end{tabular}}
\vspace{-2mm}
\caption{Comparison for known actions. Our \name{} achieves the highest top-1 accuracy, the lowest F1 and cMAP scores on two benchmarks. It is worth noting that our analysis excludes the downsampling-based methods.
}
\vspace{-2mm}
\label{tab:sota_known_actions}
\end{table}

\begin{table}
\setlength{\tabcolsep}{2.5pt}
\centering
\resizebox{0.95\columnwidth}{!}{
\begin{tabular}{cccccccccc} 
\toprule
\multirow{2}{*}{Method} & \multicolumn{3}{c}{VP-UCF101$\rightarrow$VP-HMDB51} & & \multicolumn{3}{c}{VP-HMDB51$\rightarrow$VP-UCF101}\\ 
\cmidrule{2-4}\cmidrule{6-8}
& Top-1 ($\uparrow$) & F1 ($\downarrow$)   & cMAP ($\downarrow$) & & Top-1 ($\uparrow$)  & F1 ($\downarrow$) & cMAP ($\downarrow$)\\ 
\midrule
Raw data & 51.44 & 0.673 & 75.58  &&84.20 & 0.684 & 76.62\\
\midrule
Collective \cite{icdm19} &45.93 &0.633 &73.36 &&77.12 &0.674 &76.01 \\
VITA \cite{wu_tpami} &46.78 &0.621 &73.47 &&77.48 &0.669 &76.02\\
SPAct \cite{dave2022spact} &47.81 &0.618 &72.56 &&78.13 &0.661 &75.98\\
\midrule
\textbf{Ours} &\textbf{49.56} &\textbf{0.595} &\textbf{71.25} &&\textbf{81.04} &\textbf{0.645} &\textbf{74.60} \\
\bottomrule
\end{tabular}}
\vspace{-2mm}
\caption{Comparison for unseen actions. Despite the challenging nature of this protocol, \name{} exhibits the best action-privacy trade-off amongst all evaluated methods.}
\vspace{-5mm}
\label{tab:sota_novel_actions}
\end{table}

\subsection{Implementation details}
Our \name{} is implemented using PyTorch based on a ViT-S \cite{vit, rao2021dynamicvit, deit} pre-trained on ImageNet \cite{imagenet}. Both the action recognizer and video-privacy recognizer are ViT-S-based classifiers. The optimization is carried out using AdamW \cite{loshchilov2017decoupled} with a weight decay of 0.05. The video and \tub{} size are set at {\small$16\times112\times112\times3$} and {\small$2\times16\times16\times3$}, respectively. The frame sampling rates on VP-HMDB51 and VP-UCF101 are set at 2 and 4, respectively. 
The learning rate is linearly scaled \wrt the input batch size using 
$0.001\times \frac {batch\ size}{512}$, which is decreased by cosine annealing. The experiments are conducted on NVIDIA TESLA V100 GPUs. The number of training epochs for the three phases are 80, 40, and 80, respectively. For more details, please refer to the attached code, which will also be released later. 

\begin{figure*}[!t]
\setlength\abovecaptionskip{2mm}
\centering
\begin{subfigure}{0.485\textwidth}
    \centering
    \includegraphics[width=1.0\columnwidth]
    {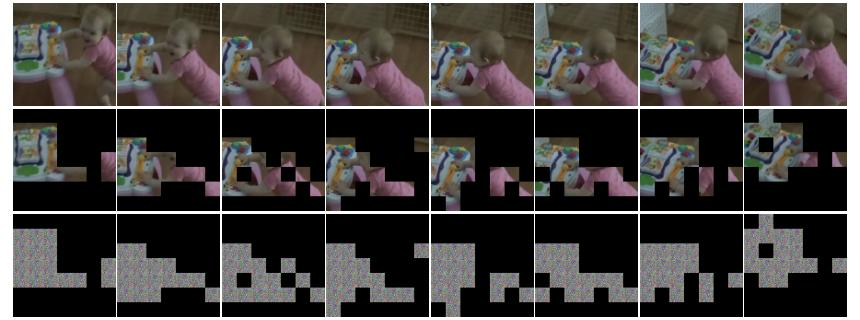}
\end{subfigure}
\hspace{-10pt}
\begin{subfigure}{0.485\textwidth}
    \centering
    \includegraphics[width=1.0\columnwidth]
    {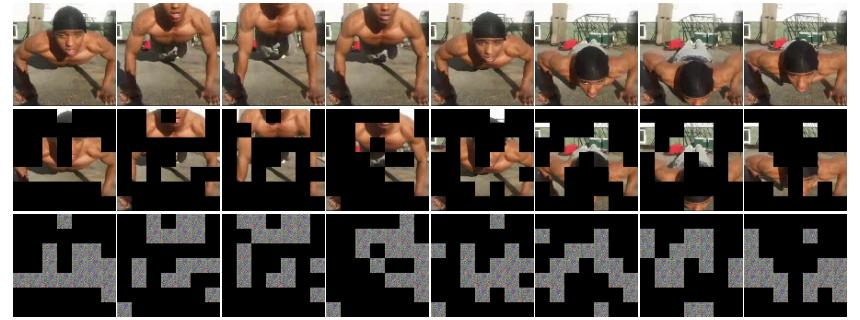}
\end{subfigure}
\begin{subfigure}{0.485\textwidth}
    \centering
    \includegraphics[width=1.0\columnwidth]
    {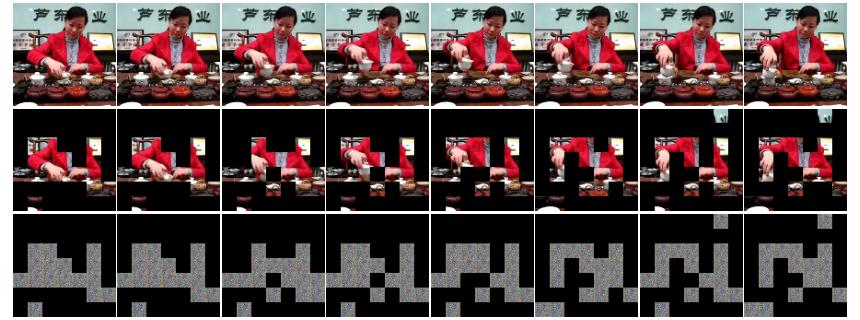}
\end{subfigure}
\hspace{-10pt}
\begin{subfigure}{0.485\textwidth}
    \centering
    \includegraphics[width=1.0\columnwidth]{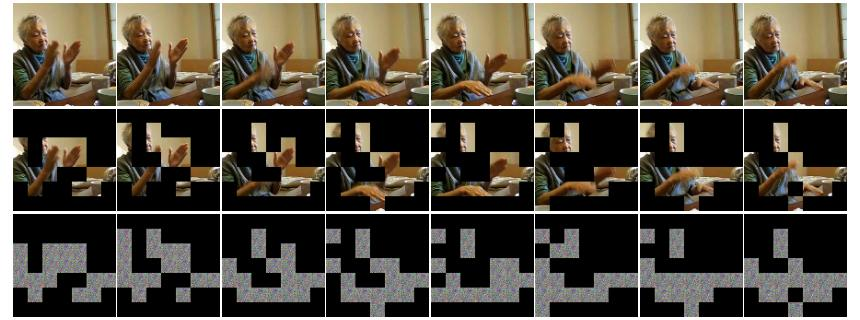}
\end{subfigure}
\caption{Visualization of our effectiveness on various actions. Each group comprises a raw video, its corresponding privacy \spars{} result, and privacy \anony{} result. The sequential actions of \textit{push}, \textit{pushup}, \textit{pour}, and \textit{clap} are arranged in a left-right top-bottom order. The compelling visual results showcase the effectiveness of our framework.} 
\vspace{-1mm}
\label{fig:qualitative_analysis}
\end{figure*}

\subsection{Comparison with state-of-the-art methods} \label{sec:sota}
To ensure fair comparisons, the same action recognizer and video-privacy recognizer as ours are implemented for evaluating SOTA methods. The results of \textit{raw data} are obtained by performing evaluations on original videos. A method exhibiting a higher top-1 accuracy alongside lower F1 and cMAP scores is deemed superior. Note that the results of Downsample-$2\times$ and $4\times$ are provided as references and excluded from comparisons. Because they merely reduce video resolutions without regard for action recognition, resulting in a dramatic decrease in the performance.

\bfsection{Comparison for known actions}
In this protocol, \name{} is trained (\ie the first two phases) and evaluated (\ie the last phase) on the same benchmark. Thus, the actions of evaluating videos are known for \name{}. The results are reported in \tab \ref{tab:sota_known_actions}. Notably, our framework exhibits the best action-privacy performance trade-off among all evaluated methods. Specifically, in comparison to SPAct on VP-HMDB51, ours achieves a top-1 accuracy that is 2.17\% higher, while demonstrating a 0.029 lower F1 score and a 1.3\% lower cMAP score. Similarly, in comparison to VITA on VP-UCF101, ours demonstrates a superior trade-off, with a 4.06\% higher accuracy, a 0.023 lower F1 score, and a 1.57\% lower cMAP score. In summary, our \name{} demonstrates the smallest action performance degradation while achieving the largest privacy performance decrease on the basis of the raw data results.

\bfsection{Comparison for unseen actions}
In this protocol, \name{} is trained and evaluated on different benchmarks. Thus, the actions of evaluating videos are unseen for \name{}. The results are reported in \tab \ref{tab:sota_novel_actions}. Note that only learning-based methods show changes in performances, compared to the experiments for known actions. Specifically, their action recognition accuracy on VP-HMDB51 decreases due to the different action sets between training and evaluating. However, their F1 and cMAP scores also exhibit apparently decreases. This could be attributed to the fact that VP-UCF101, including 101 human actions, covers a wider range of scenes of life than VP-HMDB51. Therefore, the methods trained on the former are more adaptable on privacy removal and offer better protection of privacy on the latter. This is also evidenced by the consistently degraded action-privacy trade-offs on VP-UCF101. Despite these challenges, our \name{} still achieves the best performance trade-offs on both benchmarks.

\subsection{Qualitative analysis} \label{sec:qualitative_analysis}
To qualitatively analyze our effectiveness, we provide a visualization of raw videos, the privacy \spars{} effect (as described in \sec \ref{sec:tube_sparse}), and the privacy \anony{} effect (as described in \sec \ref{sec:tub_anony}) in in \fig \ref{fig:qualitative_analysis} (more visualizations are provided in the supplementary material).

\bfsection{Sparsification reasonably abandons private \tubs{} for privacy removal} 
From the second row of each group, we observe that the \tubs{} containing private information, \eg head and face, that are not essential to action dynamics are abandoned, while those highly related to representing actions are retained. For instance, in the \textit{pour} video, the head \tubs{} are abandoned to prevent the leakage of \textit{face} and \textit{gender} information, whereas the arms are retained to represent the dynamics of pouring. Similar effects are observed in the videos of \textit{push}, \textit{pushup}, and other actions. These results demonstrate that our \name{} is capable of distinguishing private action-irrelevant \tubs{} and achieving adaptive \tub{} disentanglement for various actions. Notably, the background \tubs{} that are irrelevant to either privacy or actions are not of concern in our approach.

\begin{figure*}[!t]
\centering
\includegraphics[width=0.95\textwidth]
{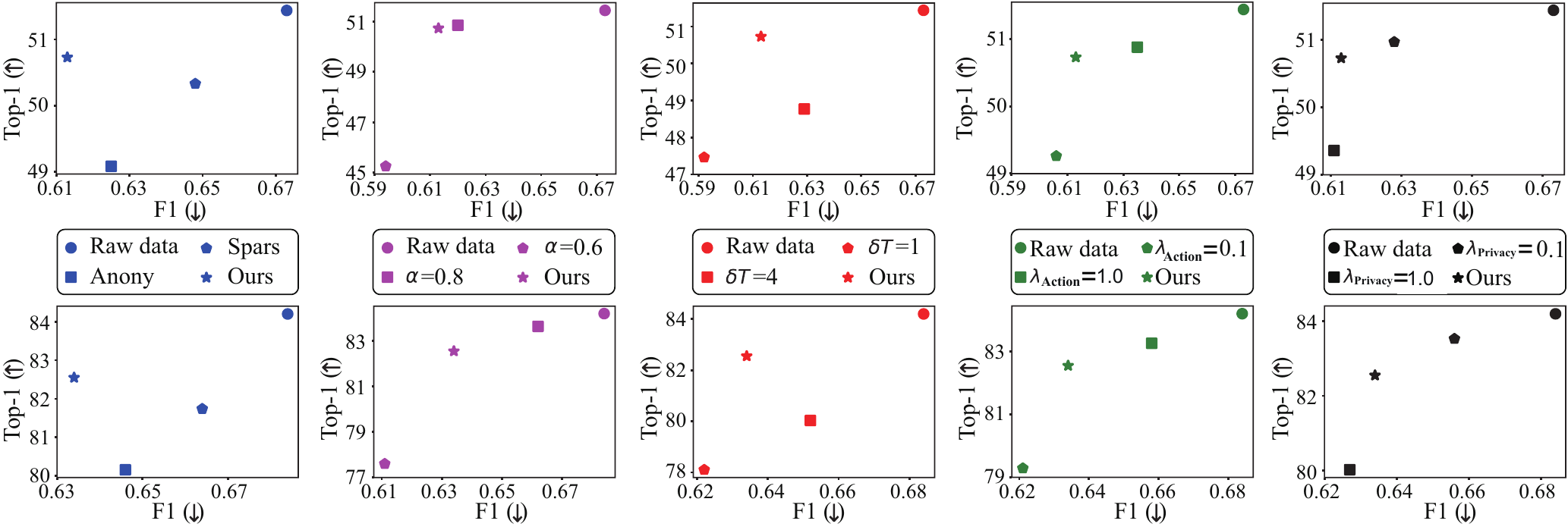}
\vspace{-3mm}
\caption{Results of the ablation experiments for separate \spars{} (sparse) and \anony{} (anony) mechanisms (1st column), \tub{} keeping proportion $\alpha$ (2nd column), temporal length of \tubs{} {\small$\delta T$} (3rd column), $\lambda_{\rm Action}$ (4th column), and $\lambda_{\rm Privacy}$ (last column) on VP-HMDB51 (1st row) and VP-UCF101 (2nd row) datasets. The experiment with the data point closest to the upper-left corner achieves the best action-privacy recognition trade-off in each chart.
}
\vspace{-5mm}
\label{fig:ablation_study}
\end{figure*}

\bfsection{Anonymization renders action \tubs{} unidentifiable for human eyes} \label{sec:complete} 
The third row of each group shows that the anonymized action \tubs{} are devoid of object silhouettes, which is a marked improvement over previous learning-based methods that rely on CNN-based transformation networks, such as ITNet \cite{johnson2016perceptual} in VITA \cite{wu_tpami} and UNet \cite{ronneberger2015u} in SPAct \cite{dave2022spact}. These methods inherently maintain spatial information of objects, and in VITA \cite{wu_tpami}, for instance, salient object silhouettes are easily recognizable in transformed frames. In contrast, our STPrivacy is based on the self-attention interactions among tokens and possesses inherent advantages in concealing geometric information of objects. This empowers our framework with a notable advantage over previous approaches in achieving high-quality visual privacy preservation.

\subsection{Ablation study} \label{sec:ablation}
In this section, we investigate the main components and hyperparameters of our \name{} through extensive ablation experiments. Each experiment involves adjusting only the specified factor while keeping other factors at the default setting of \textit{ours}. The chart results are provided in \fig \ref{fig:ablation_study}.

\bfsection{Separate \spars{} effectively reduces privacy leakage}
We perform separate training of the \spars{} mechanism by ablating the PAB of our framework. Our findings, from the results of the downsampling-based methods in \tab \ref{tab:sota_known_actions}, reveal that indiscriminate video transformations such as decreasing resolutions pose greater harm to action recognition than to privacy recognition. This disparity may be attributed to the fact that privacy recognition does not necessitate the continuity of object dynamics, as is required for action recognition. Upon analyzing the 1st column charts, our \textit{spars} experiment results in a considerably smaller decrease ratio in the top-1 accuracy when compared to the F1 and cMAP scores based on the \textit{raw data} results. This validates the effectiveness of our \spars{} mechanism in abandoning private action-irrelevant \tubs{}, thereby reducing privacy leakage.

\bfsection{Separate \anony{} successfully maintains action clues}
We also perform separate training of the \anony{} mechanism by setting $\alpha=1.0$ in the PSBs of our framework to investigate its impact on action recognition. As shown in the 1st column charts, the \textit{anony} experiment results in remarkably lower F1 and cMAP scores, but a relatively higher top-1 accuracy. It demonstrates the successful maintenance of action clues by our \anony{} mechanism when removing privacy in the embedding space.

\bfsection{Tubelet keeping proportion easily balances action privacy recognition}
We study the impact of the \tub{} keeping proportion $\alpha$ in the 2nd column charts. We find that a larger $\alpha$, encouraging \name{} to keep more \tubs{} through \equ \ref{equ:loss_ratio}, benefits action recognition while worsening privacy leakage, and vice versa. This consistent trend is observed across both benchmarks, indicating that users can easily balance the trade-off between action and privacy recognition by choosing an appropriate $\alpha$ for training \name{}. Moreover, users can choose a different $\alpha$ for deployment, allowing them to adjust the performance trade-off for inference without the need of retraining. This is achieved when selecting $\alpha^3$ of all \tubs{} for a video by sorting their keeping probabilities, as described in \sec \ref{sec:tube_sparse}. This advantage distinguishes our approach from previous learning-based methods, where the action-privacy trade-offs are frozen after training.

\bfsection{Temporal length affects the selection flexibility of remaining \tubs{}}
We adjust the temporal length of \tubs{} {\small$\delta T$} in the 3rd column charts to study its optimal setting. When {\small$\delta T$} is set to 1, each \tub{} only covers a single frame, resulting in more selections for remaining \tubs{} compared to {\small$\delta T=2$}. Consequently, the privacy removal becomes more flexible and easier. However, this also makes it more difficult to model temporal correlations among \tubs{}, which is reflected by the decreased action recognition performance. On the other hand, when {\small$\delta T$} is set to 4, the number of selections of remaining \tubs{} is considerably reduced, limiting the privacy-removal flexibility. Moreover, the length of \tubs{} becomes too long, leading to abandoned \tubs{} inevitably containing action clues, which harms action recognition. Based on these observations, we set {\small$\delta T$} to 2 as our final setting in the experiments.

\bfsection{Framework robustness in terms of the task weighting coefficients}
We choose different values for $\lambda_{\rm Action}$ and $\lambda_{\rm Privacy}$ to optimize our \name{} in the last two column charts. A higher value highlights the corresponding task during the transformation of raw videos. The small resulting fluctuations of the action-privacy trade-offs demonstrate the satisfactory robustness of our framework.

\subsection{Frame-level privacy preservation} \label{sec:frame-level-test}
Although it is originally designed for the video-level PPAR, we also conduct experiments to verify the efficacy of our \name{} on frame-level privacy preservation. Previous works evaluate their methods on PA-HMDB \cite{wu_tpami}, which comprises 515 videos with action and privacy labels. We randomly divide 60\% of the videos for training and use the remaining 40\% for testing. As per \cite{wu_tpami, dave2022spact}, we train an action recognizer and a frame-privacy recognizer from scratch on the transformed HMDB51 dataset and the transformed training set of PA-HMDB, respectively, before testing them on the transformed testing set of PA-HMDB. All methods follow the same protocol, and the results are reported in \tab \ref{tab:frame_level}. The comparison clearly demonstrates that our \name{} is still superior in this scenario. Therefore, we conclude that it is capable of protecting both video-level and frame-level privacy. 

\begin{table}
\setlength{\tabcolsep}{2pt}
\centering
\resizebox{0.55\columnwidth}{!}{
\begin{tabular}{cccc} 
\toprule
Method & Top-1 ($\uparrow$) & F1 ($\downarrow$)   & cMAP ($\downarrow$)\\ 
\midrule
Raw data &51.23 &0.572 &71.12\\
\midrule
Downsample-$2\times$ &40.38 &0.519 &67.81 \\
Downsample-$4\times$ &31.24 &0.511 &67.14\\
\midrule
Blackening \cite{dave2022spact} &38.09 &0.557 &70.02 \\
StrongBlur \cite{dave2022spact} &40.64 &0.560 &70.31 \\
WeakBlur \cite{dave2022spact} &46.89 &0.567 &70.74 \\
Collective \cite{icdm19} &46.47 &0.554 &69.96 \\
VITA \cite{wu_tpami} &47.78 &0.549 &69.45 \\
SPAct \cite{dave2022spact} &48.33 &0.543 &69.44 \\ 
\midrule
\textbf{Ours} &\textbf{50.61} & \textbf{0.523} & \textbf{68.76}\\
\bottomrule
\end{tabular}}
\vspace{-2mm}
\caption{Comparison for frame-level privacy preservation. Despite not being specifically proposed for this scenario, our \name{} exhibits substantial improvements over the existing methods.}
\vspace{-5mm}
\label{tab:frame_level}
\end{table}

\subsection{Generalization ability}
To verify the generalization ability of our \name{}, we conduct additional experiments on two related tasks. Unless specified otherwise, the training details used in these experiments are identical to those used in the experiments on our PPAR benchmarks.

\bfsection{Facial attribute-preserving dynamic expression recognition}
We term this task as FAPDER, which aims to perform dynamic expression recognition on facial videos \cite{liu2014learning, verma2019learnet} while preventing appearance attributes from leaking. It is based on the \textbf{CelebVHQ} dataset \cite{zhu2022celebvhq}, which consists of 35,666 facial videos collected from 15,653 individuals. Each video is annotated with one of eight expressions: neutral, happy, sad, anger, fear, surprise, contempt, and disgust, along with appearance attributes. We select five attributes, namely young, pointy nose, male, rosy cheeks, and brown hair, to protect. We randomly split 60\% of all videos for training and use the remaining 40\% for testing, with a frame sampling rate of 2. The results of all methods are reported in \tab \ref{tab:celebvhq}, which clearly demonstrates the superior generalization ability of our \name{} on FAPDER.

\begin{table}
\setlength{\tabcolsep}{2.2pt}
\centering
\resizebox{0.55\columnwidth}{!}{
\begin{tabular}{cccc} 
\toprule
Method & Top-1 ($\uparrow$) & F1 ($\downarrow$) & cMAP ($\downarrow$)\\ 
\midrule
Raw data &73.13 &0.714 &78.51 \\
\midrule
Downsample-$2\times$ &54.53 &0.259 &54.33 \\
Downsample-$4\times$ &41.67 &0.244 &47.17 \\
\midrule
StrongBlur \cite{dave2022spact} &56.77 &0.331 &63.61 \\
WeakBlur \cite{dave2022spact} &64.22 &0.369 &64.19 \\
Collective \cite{icdm19} &63.65 &0.348 &63.20\\
VITA \cite{wu_tpami} &66.43 &0.314 &61.09 \\
SPAct \cite{dave2022spact} &67.12 &0.326 &62.78\\ 
\midrule
\textbf{Ours} &\textbf{71.37}  &\textbf{0.280} & \textbf{59.30}\\
\bottomrule
\end{tabular}}
\vspace{-2mm}
\caption{Comparison for FAPDER. Our \name{} shows significant superiority over other methods.}
\vspace{-1.5mm}
\label{tab:celebvhq}
\end{table}

\begin{table}
\setlength{\tabcolsep}{1.5pt}
\centering
\resizebox{0.64\columnwidth}{!}{
\begin{tabular}{cccc} 
\toprule
Method & Action ($\uparrow$) & Object ($\downarrow$)  & Scene ($\downarrow$)\\ 
\midrule
Raw data &27.28 &15.72 &30.18\\
\midrule
Downsample-$2\times$ &17.08 &11.74 &24.77 \\
Downsample-$4\times$ &15.23 &10.66 &22.40 \\
\midrule
Blackening \cite{dave2022spact} &18.51 &14.89 &29.59 \\
StrongBlur \cite{dave2022spact} &19.08 &15.01 &29.32 \\
WeakBlur \cite{dave2022spact} &21.67 &14.37 &28.68 \\
Collective \cite{icdm19} &22.20 &14.27 &28.79 \\
VITA \cite{wu_tpami} &22.87 &14.44 &27.76 \\
SPAct \cite{dave2022spact} &23.16 &14.02 &28.59 \\ 
\midrule
\textbf{Ours} &\textbf{26.42} &\textbf{12.38} &\textbf{25.64}\\
\bottomrule
\end{tabular}}
\vspace{-2mm}
\caption{Comparison for OSPAR. Our \name{} outperforms the existing methods by remarkable margins.}
\vspace{-5mm}
\label{tab:phvu}
\end{table}

\bfsection{Object- and scene-preserving action recognition}
We term this task as OSPAR, which aims to recognize actions in videos while preventing object and scene categories from leaking \cite{dave2022spact}. The dataset used for this task is \textbf{P-HVU} \cite{dave2022spact}, which contains 245,212 videos for training and 16,012 videos for testing. Each video is assigned multi-label action, object and scene annotations. Thus, cMAP is applied to evaluate the three tasks. The frame sampling rate is set at 2. The results of all methods are reported in \tab \ref{tab:phvu}. The comparison reveals that our \name{} also exhibits superior generalization ability on OSPAR.

\section{Conclusion}
In this paper, we propose a novel video-level PPAR framework \name{}, which benefits action recognition and offers more strict privacy preservation. It exhibits significant advantages over existing methods. We also construct the first two large-scale benchmarks, validating our SOTA action-privacy trade-offs and qualitative effectiveness. Finally, we validate our superior generalization ability on two related tasks. These demonstrate that our contributions have the great potential to advance the PPAR research.

{\small
\bibliographystyle{ieee_fullname}
\bibliography{references}
}

\newpage
\newcommand{\hbAppendixPrefix}{}
\renewcommand{\thesection}{\hbAppendixPrefix\arabic{section}}
\setcounter{section}{0}
\renewcommand{\thefigure}{\hbAppendixPrefix\arabic{figure}}
\renewcommand{\thetable}{\hbAppendixPrefix\arabic{table}} 
\renewcommand{\theequation}{\hbAppendixPrefix\arabic{equation}} 

\onecolumn
\begin{center}
\centering
\Large  

\textbf{\name{}: Spatio-Temporal Privacy-Preserving Action Recognition}

\vspace{3 mm}
\textsc{Appendix}

\end{center}
\vspace{7.5 mm}

\section{More visualizations}
\begin{figure}[!t]
\setlength\abovecaptionskip{2mm}
\centering
\begin{subfigure}{0.8\textwidth}
    \centering
    \includegraphics[width=1.0\columnwidth]{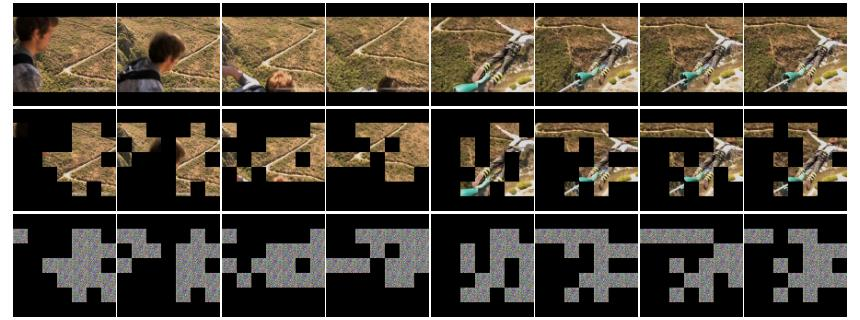}
\end{subfigure}
\hspace{-10pt}
\begin{subfigure}{0.8\textwidth}
    \centering
    \includegraphics[width=1.0\columnwidth]
    {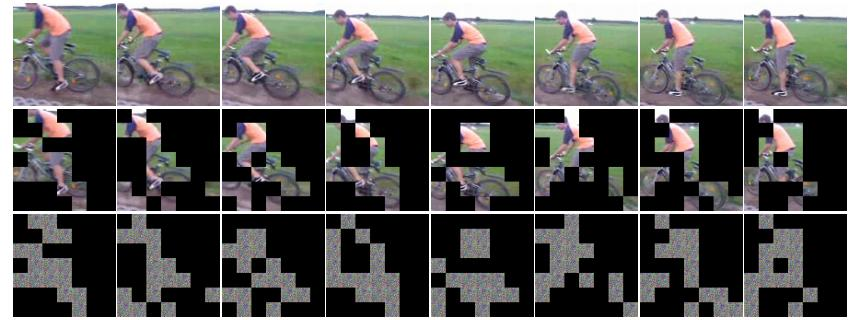}
\end{subfigure}
\begin{subfigure}{0.8\textwidth}
    \centering
    \includegraphics[width=1.0\columnwidth]{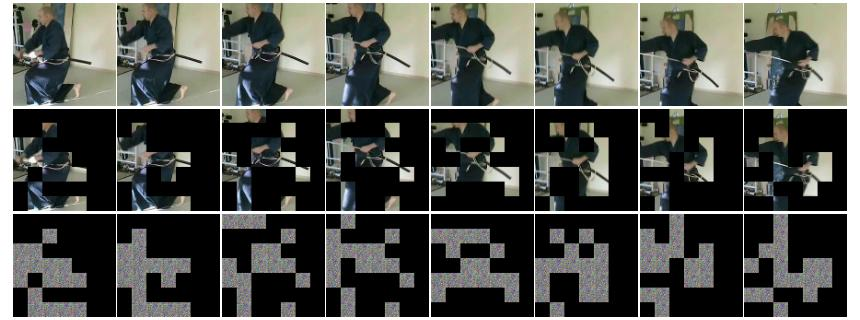}
\end{subfigure}
\hspace{-10pt}
\begin{subfigure}{0.8\textwidth}
    \centering
    \includegraphics[width=1.0\columnwidth]{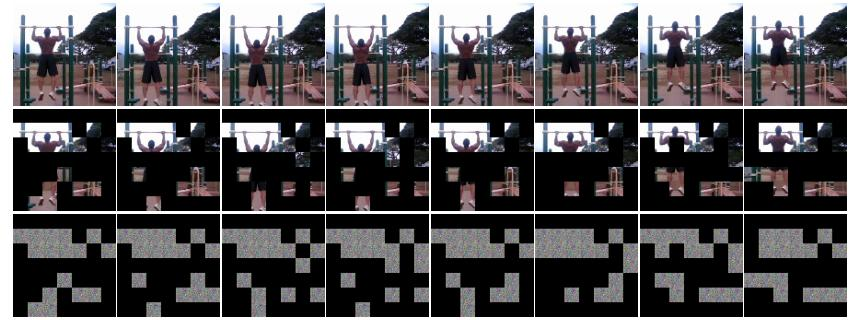}
\end{subfigure}
\caption{Visualization of our effectiveness on various actions. Each group comprises a raw video, its corresponding privacy \spars{} result, and privacy \anony{} result. The sequential actions of \textit{dive}, \textit{ride bike}, \textit{draw sword}, and \textit{pullup} are arranged in a top-bottom order.}
\vspace{-5mm}
\label{fig:more_vis_1}
\end{figure}

\begin{figure}[!t]
\setlength\abovecaptionskip{2mm}
\centering
\begin{subfigure}{0.8\textwidth}
    \centering
    \includegraphics[width=1.0\columnwidth]{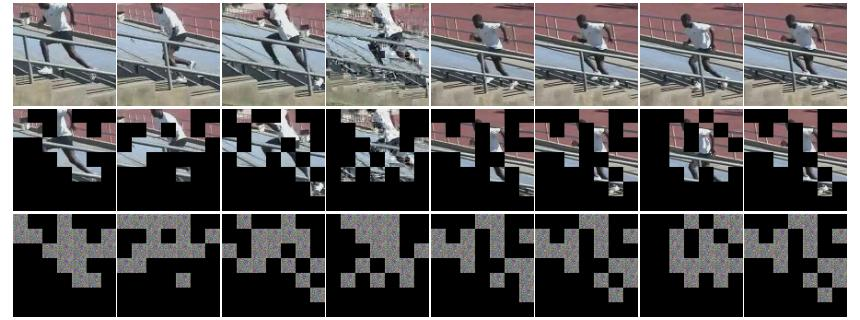}
\end{subfigure}
\hspace{-10pt}
\begin{subfigure}{0.8\textwidth}
    \centering
    \includegraphics[width=1.0\columnwidth]{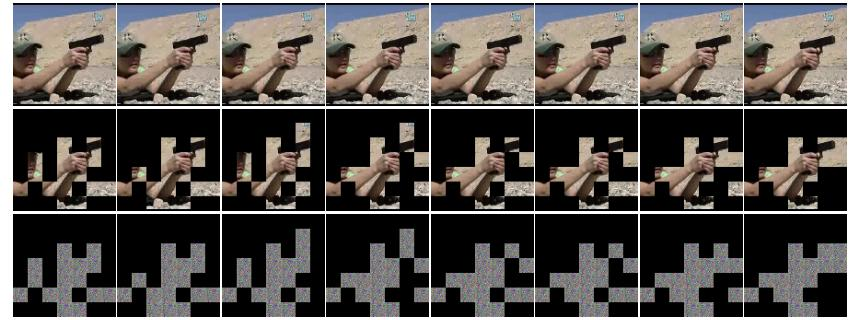}
\end{subfigure}
\begin{subfigure}{0.8\textwidth}
    \centering
    \includegraphics[width=1.0\columnwidth]
    {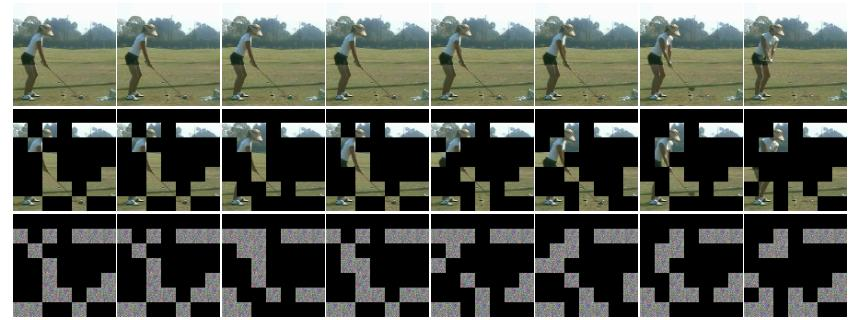}
\end{subfigure}
\hspace{-10pt}
\begin{subfigure}{0.8\textwidth}
    \centering
    \includegraphics[width=1.0\columnwidth]{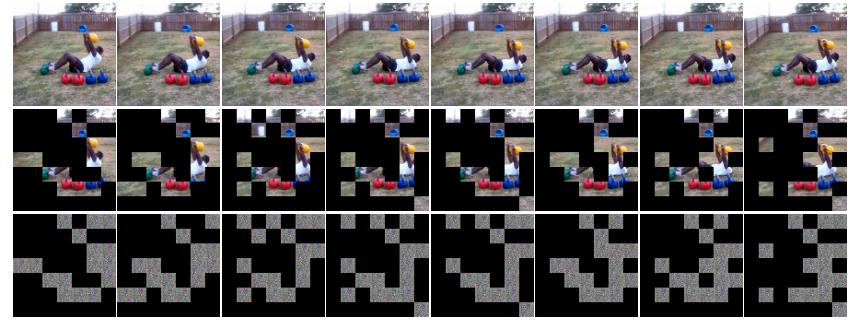}
\end{subfigure}
\caption{Visualization of our effectiveness on various actions. Each group comprises a raw video, its corresponding privacy \spars{} result, and privacy \anony{} result. The sequential actions of \textit{climb stairs}, \textit{shoot gun}, \textit{golf}, and \textit{situp} are arranged in a top-bottom order.}
\vspace{-5mm}
\label{fig:more_vis_2}
\end{figure}

\begin{figure}[!t]
\setlength\abovecaptionskip{2mm}
\centering
\begin{subfigure}{0.8\textwidth}
    \centering
    \includegraphics[width=1.0\columnwidth]{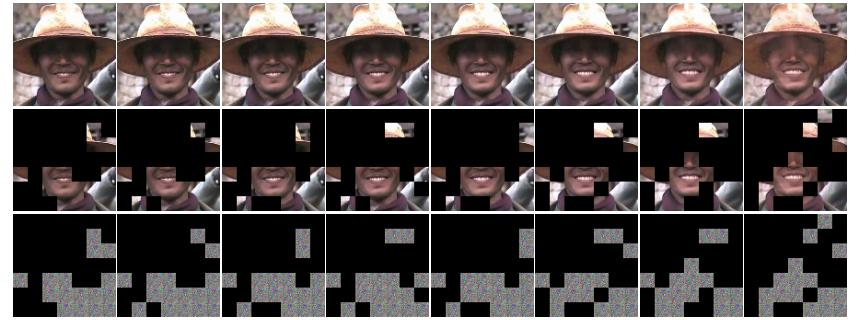}
\end{subfigure}
\hspace{-10pt}
\begin{subfigure}{0.8\textwidth}
    \centering
    \includegraphics[width=1.0\columnwidth]{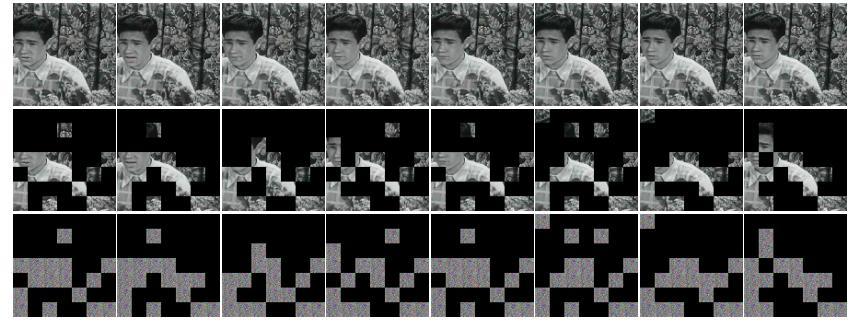}
\end{subfigure}
\begin{subfigure}{0.8\textwidth}
    \centering
    \includegraphics[width=1.0\columnwidth]
    {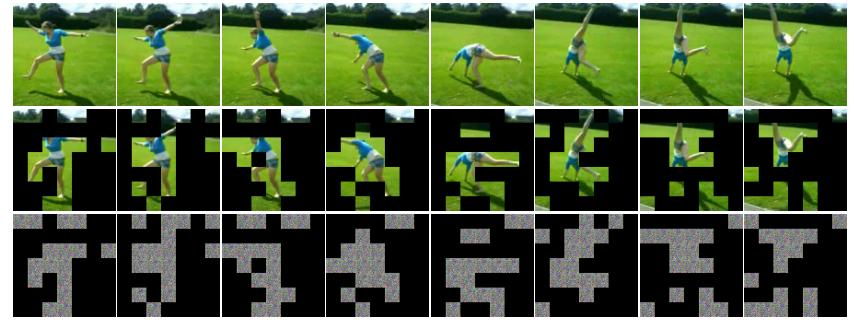}
\end{subfigure}
\hspace{-10pt}
\begin{subfigure}{0.8\textwidth}
    \centering
    \includegraphics[width=1.0\columnwidth]{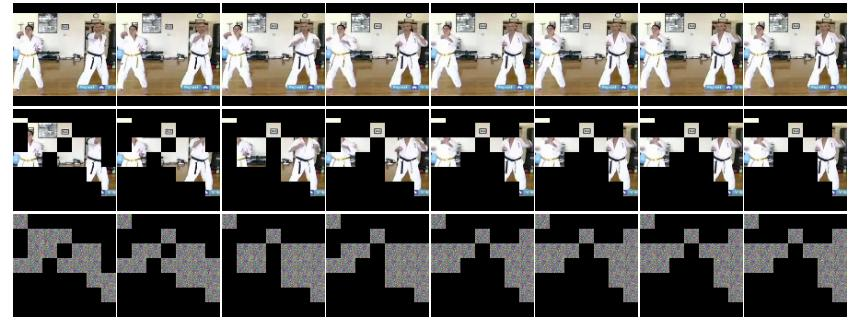}
\end{subfigure}
\caption{Visualization of our effectiveness on various actions. Each group comprises a raw video, its corresponding privacy \spars{} result, and privacy \anony{} result. The sequential actions of \textit{smile}, \textit{talk}, \textit{cartwheel}, and \textit{punch} are arranged in a top-bottom order.}
\vspace{-5mm}
\label{fig:more_vis_3}
\end{figure}

Due to the space limitation, we only show the visual effectiveness of our \name{} on a couple of actions in Section \ref{sec:qualitative_analysis}. In this appendix, we provide more visualizations in \fig \ref{fig:more_vis_1}, \ref{fig:more_vis_2}, and \ref{fig:more_vis_3}. The compelling results demonstrate the effectiveness of our framework.

\end{document}